\documentclass[5p]{elsarticle} 

\usepackage{amssymb}
\usepackage{float}
\usepackage{xcolor}
\usepackage{array}
\usepackage{booktabs}
\usepackage{bbding}

\usepackage{lineno}
\usepackage{multirow}
\usepackage{amsmath,amssymb,amsfonts}
\usepackage{algorithmic}
\usepackage{graphicx}
\usepackage{textcomp}
\usepackage{color}
\usepackage{hyperref}
\journal{Journal of Parallel and Distributed Computing}

\usepackage{ecrc}
\volume{00}
\firstpage{1}
\runauth{}
\jid{procs}
\jnltitlelogo{JPDC}
\CopyrightLine{2024}{Published by Elsevier Ltd.}

\newcommand{\colorbibs}[2][blue]%
{%
	\DeclareBibliographyCategory{ColoredBiblist#1}%
	\addtocategory{ColoredBiblist#1}{#2}%
	\AtEveryBibitem{\ifcategory{ColoredBiblist#1}{\color{#1}\bfseries}{}}
}

\begin{document}

\begin{frontmatter}



\title{Efficient Long-distance Latent Relation-aware Graph Neural Network for Multi-modal Emotion Recognition in Conversations}


\author[address1]{Yuntao Shou}
\ead{shouyuntao@stu.xjtu.edu.cn}

\author[address1]{Wei Ai}
\ead{aiwei@hnu.edu.cn}

\author[address1]{Jiayi Du}
\ead{dujiayi@csuft.edu.cn}


\author[address1]{Tao Meng\corref{cor1}}
\ead{mengtao@hnu.edu.cn}
\cortext[cor1]{Corresponding author}



\author[address3]{Haiyan Liu}
\ead{liuhy@ucmerced.edu}

\author[address2]{Nan Yin}
\ead{nan.yin@mbzuai.ac.ae}

\address[address1]{College of Computer and Mathematics, Central South University of Forestry and Technology, ChangSha, Hunan 410004, China}
\address[address3]{College of Information Engineering, Changsha Medical University, ChangSha, Hunan 410203, China}

\address[address2]{Mohamed bin Zayed University of Artificial Intelligence, UAE.}

\begin{abstract}
The task of multi-modal emotion recognition in conversation (MERC) aims to analyze the genuine emotional state of each utterance based on the multi-modal information in the conversation, which is crucial for conversation understanding. Existing methods focus on using graph neural networks (GNN) to model conversational relationships and capture contextual latent semantic relationships. However, due to the complexity of GNN, existing methods cannot efficiently capture the potential dependencies between long-distance utterances, which limits the performance of MERC. In this paper, we propose an Efficient Long-distance Latent Relation-aware Graph Neural Network (ELR-GNN) for multi-modal emotion recognition in conversations. Specifically, we first use pre-extracted text, video and audio features as input to Bi-LSTM to capture contextual semantic information and obtain low-level utterance features. Then, we use low-level utterance features to construct a conversational emotion interaction graph. To efficiently capture the potential dependencies between long-distance utterances, we use the dilated generalized forward push algorithm to precompute the emotional propagation between global utterances and design an emotional relation-aware operator to capture the potential semantic associations between different utterances. Furthermore, we combine early fusion and adaptive late fusion mechanisms to fuse latent dependency information between speaker relationship information and context. Finally, we obtain high-level discourse features and feed them into MLP for emotion prediction. Extensive experimental results show that ELR-GNN achieves state-of-the-art performance on the benchmark datasets IEMOCAP and MELD, with running times reduced by 52\% and 35\%, respectively. In addition, ELR-GNN can effectively improve the accuracy of the MERC task by capturing and fusing the latent semantic relationships between utterances.
\end{abstract}


\end{frontmatter}

\section{Introduction}
Multi-modal emotion recognition in conversations (ME-RC) has received research attention \cite{lian2022smin, shou2022conversational, shou2023low, meng2023deep, meng2024multi, ai2024gcn, shou2023adversarial, shou2024revisiting, ai2023two, meng2024revisiting} due to its wide application in the fields of intelligent customer service and emotion analysis \cite{priyasad2020attention}, human-computer interaction (HCI) \cite{zhang2020emotion}, and security monitoring \cite{chudasama2022m2fnet}. For instance, in HCI, MERC can help computers better understand the emotional state of human users, thereby enabling more intelligent interactions and improving user experience. Unlike traditional non-conversational or unimodal emotion recognition tasks \cite{li2020exploring}, MERC requires identifying the speaker's genuine emotions based on textual, auditory, and visual information in the conversation utterances \cite{shou2022object}.


The current mainstream research methods mainly use RNN, Transformer, and GCN to model conversation context and multi-modal information in MERC. For example, DialogueRNN \cite{majumder2019dialoguernn} uses a sequential approach to track conversation context and captures the most important emotional features through a memory mechanism. Although RNN-based methods can model the speaker's contextual information, they have limited memory ability for long-distance conversations, which limits the application of RNN in MERC tasks \cite{yang2024emotion}. To solve the above problems, the Transformer architecture \cite{ma2023transformer} is proposed to model long-distance context dependencies in MERC. For instance, CTNet \cite{lian2021ctnet} builds a Single Transformer and Cross Transformer to capture long-distance context dependencies and realize intra-module and inter-module information interaction for emotion recognition. However, methods based on Transformer architecture ignore conversational relationship information between speakers, which limits the model's emotion recognition performance \cite{li2023graphcfc, shou2023graphunet}. To tackle this limitation, many GCN methods have been proposed to model interaction information between speakers. For example, DialogueGCN \cite{ghosal2019dialoguegcn} uses a graph structure to model conversation context and uses GCN to learn conversation graphs to achieve semantic understanding and emotional recognition of conversations. In addition, LR-GCN \cite{ren2021lr} believes that the context latent dependencies of utterences should also be considered. LR-GCN uses multi-head attention to construct multiple full association graphs to model potential conversational relationships, and then uses GCN to learn latent relationships to achieve emotion recognition. However, limited by the complexity of GCN, these methods usually adopt a fixed window size strategy and then fully connect the utterances within the window to construct a conversation graph, which significantly limits the ability to obtain long-distance contextual information.

Inspired by LR-GCN, we also use GCN to model dialogue relationship information between speakers for MERC. Furthermore, long-distance context potential dependencies can provide more information for emotion classification and help reveal the genuine emotion of utterances. Therefore, how to comprehensively consider long-distance contextual dependencies while ensuring that the number of model parameters does not increase dramatically remains a challenge.

In this paper, we propose an Efficient Long-distance Latent Relation-aware Graph Neural Network (ELR-GNN) for multi-modal emotion recognition in conversation. Specifically, we first use RoBERTa, 3D-CNN, and openSMILE to perform pre-feature extraction of text, video, and audio features, respectively. Next, we use Bi-LSTM to capture contextual semantic information and obtain low-level utterance features. We then use low-level utterence features to construct a speaker graph. In the constructed speaker relationship graph, low-level utterence features are used as node features, while dialogue relationship information between speakers is used for edge construction. To capture the latent dependency information between long-distance contexts, we use the graph random neural network algorithm to randomly sample top-$k$ nodes for information extraction. In addition, we combine early fusion and adaptive late fusion mechanisms to simultaneously fuse speaker relationship information and latent dependency information between contexts. Finally, we fine-grained obtained high-level utterance features and fed them into the MLP and softmax function for emotion prediction.

\begin{itemize}
	\item We propose a novel Efficient Long-distance Latent Relation-aware Graph Neural Network (ELR-GNN) for MER. ELR-GCN not only considers conversational relationship information between speakers, but also captures long-distance context latent dependency information.
	
	\item We propose a graph random neural network architecture in which long-distance latent dependencies between utterances are captured by randomly sampling top-$k$ node features. Furthermore, we combine early fusion and adaptive late fusion mechanisms to simultaneously exploit speaker information and context's latent dependency information during information propagation.
	
	\item We perform extensive experiments on two publicly available datasets to verify the effectiveness of the ELR-GNN method.
\end{itemize}

\section{Related Work}

\subsection{Multi-modal Emotion Recognition}
Single-modality emotion recognition may be limited. For example, text-based emotion recognition alone may not capture the emotional cues in speech and facial expressions \cite{zhang2023structure}. Multimodal emotion recognition (MER) can integrate multiple information sources to improve the accuracy and robustness of emotion recognition.

Current mainstream MER research mainly focuses on RNN, Transformer and GCN. For instance, DialogueRNN \cite{majumder2019dialoguernn} modeled individual speakers and uses three different GRUs to achieve more effective correlations between speakers. DialogueGCN \cite{ghosal2019dialoguegcn} was proposed to solve the problem that RNN-based methods cannot consider the dialogue relationship between speakers. DialogueGCN improves the performance of MER by modeling the interactive relationship between speakers through the inherent properties of the graph structure and using graph convolution operations to transfer contextual semantic information. TL-ERC \cite{hazarika2019emotion} used transfer learning methods to solve problems in supervised learning that require large amounts of high-quality annotated data. CTNet \cite{lian2021ctnet} proposed a multi-modal learning framework, which achieves cross-modal contextual semantic information interaction by building a single Transformer and a cross Transformer.

However, current mainstream methods only consider contextual semantic information, latent dependencies of local utterances, and conversational relationships between speakers, and their focus is on exploring the semantic information between utterances and the correlation between speakers. The above approach ignores latent dependencies of the global context, which limits the performance of MER.

\begin{figure*}
	\centering
	\includegraphics[width=1\linewidth]{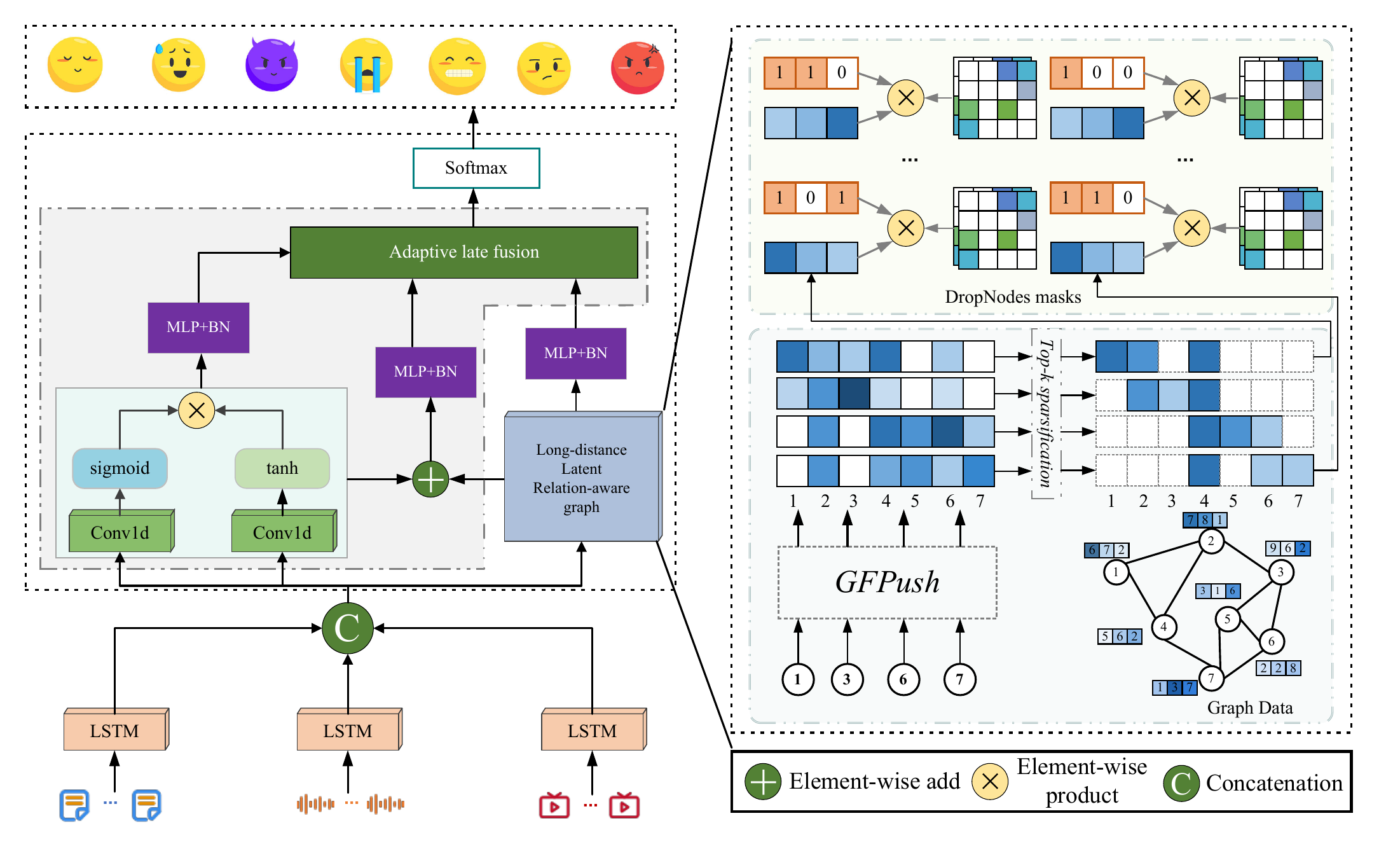}
	\caption{The overall architecture of ELR-GCN for milti-modal emotion recognition. ELR-GCN contains auxiliary information module and graph random neural network module. The auxiliary information module is used to achieve further extraction of contextual semantic information and fusion of speaker relationships and long-distance latent relationships through early and adaptive late fusion. The graph random neural network module is used to model speaker relationships and long-distance contextual latent dependencies.}
	\label{fig:architecture}
\end{figure*}

\subsection{Scalable Graph Neural Network}
The current mainstream scalable GNN methods include three types of methods: 1) Node sampling strategy: Accelerate the aggregation process of node features by sampling nodes. The representative methods include GraphSAGE \cite{hamilton2017inductive}, FastGCN \cite{chen2018fastgcn} and LADIES \cite{zou2019layer}. The main idea of GraphSage is to update the representation of each node through multiple rounds of neighbor sampling and aggregation of neighbor node information, thereby capturing the structure and relationships between nodes in graph data. FastGCN proposes a structured graph node sampling strategy, which selects sampling nodes by considering the structural information of the graph to retain important structural features of the graph. This sampling strategy can preserve the graph information as much as possible while ensuring sampling efficiency. LADIES adopts an adaptive density modeling method to capture local and global information by learning the density distribution of neighbors around a node. LADIES can effectively update the representation of nodes to one that takes into account both local and global information. 2) Graph partitioning method: Divide the original large graph into several small subgraphs and run GNN on the subgraphs. The mainstream graph partitioning methods include Cluster-GCN \cite{chiang2019cluster} and GraphSAINT \cite{zeng2019graphsaint}. The mainstream graph partitioning methods include Cluster-GCN and GraphSA-INT. Cluster-GCN divides the original large-scale graph data into multiple subgraphs, each subgraph contains a part of nodes and corresponding edges, thereby reducing computational and memory overhead. GraphSAINT processes large-scale graph data through graph sampling and iterative coarsening. 3) Matrix approximation method: Accelerate feature propagation by decoupling feature propagation and nonlinear transformation. SGC simplifies the nonlinear activation function in traditional graph convolutional networks, retaining only graph convolution operations.

\subsection{Multi-head Attention}
The multi-head attention in MER can help the model effectively capture the correlation information between different modalities and adaptively focus on the most important parts for the emotion classification task. For example, TEMMA \cite{9257201} proposesd a multi-modal multi-head attention for MER to comprehensively consider the complementarity and redundancy between modalities. TE-MMA can realize the semantic information interaction between modalities and capture the temporal dependence within the modalities. GA2MIF \cite{li2023ga2mif} constructed a multi-head directed graph attention network and a multi-head pairwise cross-modal attention network respectively to ach-ieve contextual semantic information extraction and cross-modal information fusion. EEANet \cite{yang2023deep} used a multi-head self-attention mechanism to capture the discriminative features in contextual semantic information that are most suitable for emotion classification.

We apply an attention mechanism to the utterance features obtained through graph convolution operations to calculate the correlation between contexts and capture the utterances with the strongest emotional features among the global context latent dependencies. Our method ELR-GNN can simultaneously consider contextual semantic information, interaction information between speakers, and latent dependency information of the global context.

\section{Proposed Method}
The overall framework of the ELR-GNN proposed in this paper is shown in Fig. 1. ELR-GNN consists of four stages, including sequential contextual feature extraction, graph construction, long-distance contextual latent relationship exploration, and information fusion. In the following subsections, we describe these four key parts in detail.

\subsection{Sequential context information extraction}
The speaker's emotional state is not only related to the textual semantic information at the current moment, but also related to the previous contextual semantic information. Therefore, we use Bi-LSTM to capture contextual semantic information in multi-modal features to more accurately understand the speaker's emotional changes. The formula of LSTM is defined as follows:
\begin{equation}\begin{aligned}
		&\left.\left[\begin{array}{c}\widetilde{C}_t\\O_t\\r_t\\z_t\end{array}\right.\right] \left.=\left[\begin{array}{c}\tanh\\ sigmoid \\ sigmoid\\ sigmoid \end{array}\right.\right]W_T\left[\begin{array}{c}u_i^t\\h_i^{t-1}\end{array}\right]  \\
		&\begin{aligned}C_t=C_t\odot r_t+C_{t-1}\odot z_t\end{aligned} \\
		&h_{i}^{t}=O_{t}\odot\mathrm{tanh}\left(C_{t}\right)
\end{aligned}\end{equation}
where $u_i^t$ represents the concatenated multi-modal features, $h^t_i$ represents the hidden layer state, $r_t$ represents the input gate, $z_t$ represents the forgetting gate, $C_t$ represents the cell state, and $\odot$ represents Hadamard product, $W$ is a learnable network parameter.

Bi-LSTM is composed of forward and reverse LSTM, and its formula is defined as follows:
\begin{equation}
	\tilde{h}_i^t=[\overrightarrow{h}_i^t,\overleftarrow{h}_i^t]
\end{equation}
where $\tilde{h}_i^t$ is obtained by concatenating the contextual semantic features extracted by forward and reverse LSTM.

\subsection{Graph Construction}
We use the inherent properties of the graph structure to construct a speaker relationship graph, in which the contextual semantic features extracted through Bi-LSTM are used as node features of the graph, and the dialogue relationships between speakers are used as edges. Specifically, given a speaker dialogue graph $\mathcal{G}=\{\mathcal{W}, \mathcal{V}, \mathcal{E}, \mathcal{R}\}$, where the node $v_i(v_i \in \mathcal{V})$ is composed of contextual semantic features (i.e., $\tilde{h}_i$), the edge $e_{ij}=1(e_{ij} \in \mathcal{R})$ indicates that there is a conversation relationship between node $v_i$ and node $v_j$, otherwise $e_{ij}=0$, $\omega_{ij}(\omega_{ij} \in \mathcal{W}, 0 \leq \omega_{ij} \leq 1)$ represents the weight of edge $e_{ij}$, and $r \in \mathcal{R}$ represents the edge relationship.

\subsection{Long-distance Latent Context Relationship Extraction}
Unlike previous work that set the context window size to 10 (i.e., the number of nodes), to capture long-distance latent dependencies of contexts, we adopt a larger context window to explore potential correlations between contexts. Specifically, we first construct an original graph $\mathcal{G}$ with a larger context window, the generalized forward push algorithm is then used to calculate the propagation matrix of the row vectors, and top-$k$ sparsification is used to further reduce the training time of the network, so as to comprehensively consider the latent correlation of the context.

\subsubsection{Propagation Matrix}
We use a mixed-order matrix of feature propagation to aggregate neighbor node information of different orders in the graph to obtain long-distance contextual latent dependency information. The formula of the propagation matrix is defined as follows:
\begin{equation}
	\Pi=\sum_{n=0}^Nw_n\cdot\left(\mathbf{\widetilde{\mathrm{D}}^{-1}\widetilde{\mathrm{A}}}\right)^n
\end{equation}
where $w_n \geq 0$ and $\sum_{n=0}^Nw_n = 1$, $\widetilde{\mathrm{A}}$ is the adjacency matrix, and $\widetilde{\mathrm{D}}$ is the degree matrix. The propagation matrix can fuse different orders of neighbor node information and capture important contextual potential dependency information by adjusting the weights.

Then we aggregate the node features and update the node features, defined as follows:
\begin{equation}
	\overline{\mathbf{X}}_s=\sum_{\upsilon\in\mathcal{N}_s^\pi}\mathbf{z}_{s}\cdot\Pi(s,\upsilon)\cdot h_s
	\label{eq:training}
\end{equation}
where $\mathbf{z}_{\mathcal{\upsilon}}\sim\text{Bernoulli}(1-\delta)$, $\Pi$ is the row vectors of the node $s$, $\mathcal{N}_s^\pi$ is the indices of non-zero value of $\Pi_s$. Through Eq. \ref{eq:training}, we can solve the problem of slow inference speed caused by the high computational complexity of GCN and achieve rapid training of the model. Therefore, we can construct larger graphs to capture long-distance context latent dependencies.

However, $\Pi_s$ is actually a difficult estimation problem. To address the problem, We use a two-stage estimation step for calculation, which includes Generalized Forward Push (GFP) and Top-$k$ sparsification. First, GFP gives the error bound of $\Pi_s$, and then Top-$k$ sparsification only retains top-$k$ elements to achieve faster calculation speed.

\subsubsection{Generalized Forward Push}
Since the row-normalized adjacency matrix $\mathbf{\widetilde{\mathrm{D}}^{-1}\widetilde{\mathrm{A}}}$ is also an inverse random walk transition probability matrix on $\mathcal{G}$, we design an efficient GFP estimation algorithm to estimate $\Pi_s$. The key step of GFP is to accelerate the random walk probability diffusion process through pruning operation. Specifically, we first give two initial vectors $q^{(n)} \in \mathbb{R}^{|V|}$ and $r^{(n)} \in \mathbb{R}^{|V|}$, and both $q^{(0)}$ and $r^{(0)}$ are initialized to $e^{(s)}$, where $e^{(s)}=1$ and $e^{(v)}=0$ for $s \neq v$. Furthermore, $q^{(n)}=0, r^{(n)}=0, 1\leq n\leq N$. Then, the GFP algorithm begins to iteratively update the $q^{(n)}$ and $r^{(n)}$ vectors through $\mathrm{r}_u^{(n)}\leftarrow\mathrm{r}_u^{(n)}+\mathrm{r}_\upsilon^{(n-1)}/\mathrm{d}_\upsilon$ and $q_u^{(n)}\leftarrow r_u^{(n)}$ until node $v$ satisfies $r_{\upsilon}^{(n-1)}>d_{\upsilon}\cdot r_{max}$, where $\mathbf{d}_{\mathcal{\upsilon}}=\widetilde{\mathbf{D}}(\upsilon,\upsilon)$. When the GFP iteration is complete, we get $\tilde{\Pi}_s\leftarrow\sum_{n=0}^Nw_n\cdot\mathbf{q}^{(n)}$.

\subsubsection{Top-$k$ Sparsification}
To reduce the computational complexity of GCN, we perform top-$k$ sparsification on $\tilde{\Pi}_s$ to accelerate model training. The core idea of Top-$k$ sparsification is to retain only the top-$k$ largest elements of $\Pi$, and set other elements to 0. Therefore, $\tilde{\Pi}^{(k)}$ has only $k$ non-zero elements, which preserves the most important emotion features in the latent dependencies of the context.

\subsubsection{Learnable Information Propagation}
Therefore, we introduce a learnable parameter $W$ to achieve dimensionality reduction of multi-modal features while improving the learning ability of the model. The formula is defined as follows:
\begin{equation}
	\overline{\mathrm{X}}_s=\sum_{\upsilon\in\mathcal{N}_s^{(k)}}\mathbf{z}_{v}\cdot\widetilde{\Pi}^{(k)}(s,\upsilon)\cdot h_{v}\cdot\mathbf{W}
\end{equation}

\subsection{Auxiliary Information Module}
Graph random neural networks can effectively extract dialogue relationship information between speakers and long-distance context potential dependency information, but it is easy to ignore some discriminative original full-emotion features. Therefore, we use AIM to extract and fuse higher-level emotional features, adaptively aggregating original emotional features, speaker relationship information, and long-distance context potential dependency information.

\subsubsection{Feature Extractor (AIM-FE)}
Multimodal data are characterized by noise and high dimensionality. To achieve denoising and capture discriminative emotional features in multi-modal data, we introduce gated convolutional networks to capture auxiliary information. In the gated convolutional network, we use sigmoid and tanh functions, which can retain the most important emotional feature information and improve the nonlinear fitting ability of the model. The formula of the gated convolutional network is defined as follows:
\begin{equation}
	\begin{aligned}
	\mathbf{Z}_{\mathcal{C}} &=tanh\left(Conv1D\left(\tilde{h}_i^t \right) \right) \\ & \odot sigmoid\left(Conv1D\left(\tilde{h}_i^t \right)\right)
	\end{aligned}
\end{equation}
where $Conv1D$ represents 1D convolution operations, $\odot$ represents Hadama product.

\subsubsection{Late Adaptive Fusion (AIM-LAF)}
To capture finer-grained semantic information in multi-modal data, early and late adaptive fusion mechanisms are combined to capture auxiliary information with fine-grained emotional features. Specifically, late fusion fuses highly abstract time and space information, ignoring detailed information. Therefore, the combination of early and late adaptive fusion mechanisms proposed in this paper can more effectively capture more discriminative emotional features adaptively from multi-modal data.

In the early fusion process, we map the contextual features $	\mathbf{Z}_{\mathcal{C}}$ through the gated convolutional network and the latent features $	\mathbf{Z}_{\mathcal{G}}$ through the graph random neural network to the same dimension, obtain $\tilde{\mathbf{z}_g}$ and $\tilde{\mathbf{z}_c}$ and fuse them. The formula is defined as follows:
\begin{equation}
	\tilde{\mathbf{z}_f}=\Omega(\tilde{\mathbf{z}_g},\tilde{\mathbf{z}_c})
\end{equation}
where $\Omega(\cdot)$ represents summation average operation. Then we use a FCN to achieve feature dimensionality reduction and obtain $\mathbf{z}_g$, $\mathbf{z}_c$, and $\mathbf{z}_f$. Then we use the attention mechanism to obtain the corresponding attention score as follows:
\begin{equation}
	e_g=\mathbf{q}^T\cdot\tanh(\mathbf{W}\cdot\mathbf{z}_g^T+\mathbf{b})
	\label{eq:atten}
\end{equation}
where $q$ represents the query matrix, $W$ and $b$ are the learnable parameters. Likewise, $e_c$ and $e_f$ are calculated using Eq. \ref{eq:atten}. Then we use the softmax function to normalize the attention coefficient as follows:
\begin{equation}
	\varphi_g=\frac{\exp(e_g)}{\exp(e_g)+\exp(e_c)+\exp(e_f)}
\end{equation}

Finally, we perform a weighted sum of ${\mathbf{z}_g},{\mathbf{z}_c}$ and ${\mathbf{z}_f}$ to obtain the final emotional feature vector representation. The formula is defined as follows:

\begin{equation}
	\mathbf{z}=\varphi_g\cdot\mathbf{z}_g+\varphi_c\cdot\mathbf{z}_c+\varphi_f\cdot\mathbf{z}_f
\end{equation}

\subsection{Model Training}
The final emotional feature vector $\mathbf{z}$ with contextual semantic information, dialogue relationship information between speakers, and long-distance latent dependency information is fed into the MLP with residual connections for feature conversion, and then use the softmax layer to get the probability of C-class emotion category:
\begin{equation}
	\begin{aligned}Z&=\mathbf{z}+ReLU(	\mathbf{z}W_z+b_z)\\\\P&=\text{softmax}(ZW_Z+b_Z)
\end{aligned}
\end{equation}
where $W_z$, $b_z$, $W_Z$, $b_Z$ is the learnable parameters. We then obtain the index of the maximum emotion probability by using the argmax function.
\begin{equation}
	\hat{y}^{(j)}=argmax(P^{(j)})
\end{equation}

Finally, we use cross-entropy loss to complete the optimization of the model:
\begin{equation}
	L=-\frac1{\sum_{i=1}^ML_i}\sum_{i=1}^M\sum_{j=1}^{L_i}\sum_{c=1}^Cy_{i,c}^{(j)}log_2(\hat{y}_{i,c}^{(j)})
\end{equation}
where $M$ represents the number of dialogues, and $L_i$ represents the number of utterances in the $i$-th dialogue.

\section{Experiments}

\subsection{Datasets}
We evaluate the ELR-GNN model proposed in this paper on two benchmark datasets, IEMOCAP \cite{busso2008iemocap} and MELD \cite{poria2019meld}. All these data sets contain three modal data sets: text, video, and audio.

IEMOCAP is a public dataset widely used in emotion recognition research. This dataset was created by the Sippy team at the University of Southern California and aims to provide detailed annotations of emotional interactions and speech/non-verbal behaviors. The IEMOCAP dataset emotionally annotates speech and video, including six emotion categories: happy, sad, angry, excited, frustrated, and neutral. Emotional annotation is accomplished through consistent annotation of data by multiple evaluators. The IEMOCAP dataset contains text, audio, and video data from 10 different actors. Each actor participated in a series of emotional interaction tasks.

MELD is an open multi-modal dataset for emotion analysis research. It was created by researchers at the University of Toronto to advance research into natural language and speech emotion recognition. The MELD data set contains data in three modalities: text, video and audio. The text is the script text from the movie dialogue. The MELD dataset contains annotations for six emotion categories: joy, sadness, anger, fear, surprise and neutral. These emotion annotations are performed independently by multiple annotators.

\begin{table*}[!t]
	\renewcommand\arraystretch{1.5}
	\setlength{\tabcolsep}{8.7pt}
	\caption{Comparison with other baseline models on the IEMOCAP dataset.}
	\label{tab:iemocap}
	\begin{tabular}{l|ccccccc}
		\hline
		\multirow{3}{*}{Methods} & \multicolumn{7}{c}{IEMOCAP}                                                                      \\ \cline{2-8}
		& Happy        & Sad         & Neutral     & Angry       & Excited     & Frustrated  & Average(w)  \\ \cline{2-8}
		& Acc.   F1    & Acc.   F1   & Acc.   F1   & Acc.   F1   & Acc.   F1   & Acc.   F1   & Acc.   F1   \\ \hline
		TextCNN                  & 27.7   29..8 & 57.1   53.8 & 34.3   40.1 & 61.1   52.4 & 46.1   50.0 & 62.9   55.7 & 48.9   48.1 \\
		bc-LSTM                  & 29.1   34.4  & 57.1   60.8 & 54.1   51.8 & 57.0   56.7 & 51.1   57.9 & 67.1   58.9 & 55.2   54.9 \\
		{MFN} & 24.0  34.1 & 65.6  70.5 & 55.5  52.1 & 72.3  66.8 & 64.3  62.1 & 67.9  62.5 & 60.1  59.9 \\
		CMN                      & 25.0   30.3  & 55.9   62.4 & 52.8   52.3 & 61.7   59.8 & 55.5   60.2 & \textbf{71.1} 60.6   & 56.5   56.1 \\
		LFM                      & 25.6   33.1  & 75.1   78.8 & 58.5   59.2 & 64.7   65.2 & 80.2   71.8 & 61.1   58.9 & 63.4   62.7 \\
		ICON                    & 22.2  29.9  & 58.8  64.6 & 62.8 57.4 & 64.7  63.0 & 58.9  63.4 & 67.2  60.8 & 59.1  58.5 \\
		A-DMN                    & 43.1   50.6  & 69.4   76.8 & 63.0   62.9 & 63.5   56.5 & \textbf{88.3}   77.9 & 53.3   55.7 & 64.6   64.3 \\
		{DialogueGCN}              & {40.6 42.7}    & \textbf{89.1 84.5}   & {62.0 63.5  } &{ 67.5 64.1 }  & {65.5 63.1 }  & {64.1 66.9}   & {65.2 64.1 }  \\
		{RGAT}                   & {60.1 51.6}    & {78.8 77.3}   & {60.1 65.4}   & {70.7 63.0}  & {78.0 68.0}   & {64.3 61.2}   & {65.0 65.2}   \\
		AGHMN                   & 48.3  52.1 & 68.3  73.3 & 61.6  58.4 & 57.5  61.9 & 68.1  69.7 & 67.1  62.3 & 63.5  63.5 \\
		BiERU                   & 54.2  31.5 & 80.6  84.2 & 64.7  60.2 & 67.9  65.7 & 62.8  74.1 & 61.9  61.3 & 66.1  64.7 \\
		{CoMPM}            & {59.9 60.7}    & {78.0 82.2}   & {60.4 63.0}   & {70.2 59.9}   & {85.8 78.2}   & {62.9 59.5}   & {67.7 67.2}  \\
		{EmoBERTa}                 & {56.9 56.4}   &{79.1 83.0}   & {64.0 61.5}   & {70.6 69.6}   & {86.0 78.0}   & {63.8 68.7}   & {67.3 67.3}   \\
		{COGMEN}                 &{57.4 51.9}    & {81.4 81.7}   & 65.4 \textbf{68.6}   & {69.5 66.0}   & {83.3 75.3}  &{63.8 68.2}  & {68.2 67.6}  \\
		CTNet                   & 47.9  51.3 & 78.0  79.9 & \textbf{69.0}  65.8 & \textbf{72.9}  67.2 & 85.3  78.7 & 52.2  58.8 & 68.0 67.5 \\
		{LR-GCN}                 &{54.2   55.5}  & {81.6   79.1} & {59.1 63.8}   & {69.4   69.0} & {76.3  74.0} & 68.2 \textbf{68.9}   & {68.5   68.3} \\
		DER-GCN                  & {60.7   58.8}  & 75.9 79.8   & {66.5} 61.5   & 71.3   \textbf{72.1} & 71.1   73.3 & 66.1   67.8 & {69.7   69.4} \\
		ELR-GCN   & \textbf{64.7 62.9} & 75.7 80.8  &  66.2 62.4  & 70.7 70.0   &  76.8 \textbf{78.6}  &  67.9 68.1  & \textbf{70.6 70.9}  \\ \hline
	\end{tabular}
\end{table*}

\subsection{Baselines and Evaluation Metrics}

\textbf{bc-LSTM} \cite{poria2017context} performs final emotion recognition by extracting the sequential context information of the utterance, which is context-sensitive.

\textbf{Text-CNN} \cite{kim2014convolutional} uses convolution filters to extract local semantic information from utterances, which is context-independent.

\textbf{MFN} \cite{zadeh2018memory} designs a multi-view learning mechanism to capture view-specific and cross-view semantic information, but MFN does not consider contextual information.

\textbf{CMN} \cite{hazarika2018conversational} achieves the fusion of speaker information and multi-modal features by introducing an attention mechanism.

\textbf{ICON} \cite{hazarika2018icon} uses GRU to extract contextual information of multi-modal features and uses attention layers to achieve the fusion of multi-modal semantic information.

\textbf{DialogueRNN} \cite{majumder2019dialoguernn} constructs three different gating units to achieve the extraction and fusion of speaker information, emotional information and global information.

\textbf{DialogueGCN} \cite{ghosal2019dialoguegcn} DialogueGCN constructs a speaker relationship graph by using contextual semantic features, and utilizes contextual semantic information and speaker relationship information to achieve emotion classification.

\begin{table*}[!t]
	\renewcommand\arraystretch{1.5}
	\caption{Comparison with other baseline models on the MELD dataset.}
	\label{tab:meld}
	\setlength{\tabcolsep}{5.4pt}{
		\begin{tabular}{l|cccccccc}
			\hline
			\multirow{3}{*}{Methods} & \multicolumn{8}{c}{MELD}                                                                       \\ \cline{2-9}
			& Neutral   & Surprise  & Fear      & Sadness   & Joy       & Disgust   & Anger     & Average(w) \\ \cline{2-9}
			& Acc. F1   & Acc. F1   & Acc. F1   & Acc. F1   & Acc. F1   & Acc. F1   & Acc. F1   & Acc. F1    \\ \hline
			TextCNN                  & 76.2 74.9 & 43.3 45.5 & 4.6 3.7   & 18.2 21.1 & 46.1 49.4 & 8.9 8.3   & 35.3 34.5 & 56.3 55.0  \\
			bc-LSTM                  & 78.4 73.8 & 46.8 47.7 & 3.8 5.4   & 22.4 25.1 & 51.6 51.3 & 4.3 5.2   & 36.7 38.4 & 57.5 55.9  \\
			DialogueRNN              & 72.1 73.5 & 54.4 49.4 & 1.6 1.2   & 23.9 23.8 & 52.0 50.7 & 1.5 1.7   & 41.0 41.5 & 56.1 55.9  \\
			{DialogueGCN}              & {70.3 72.1} & {42.4 41.7} & {3.0 2.8}   & {20.9 21.8} & {44.7 44.2} & {6.5 6.7}   & {39.0 36.5} & {54.9 54.7}  \\
			{RGAT}                     & {76.0 78.1} & {40.1 41.5} & {3.0 2.4}   & {32.1 30.7} & {68.1 58.6} & {4.5 2.2 }  & {40.0 44.6} & {60.3 61.1}  \\
			{CoMPM}                    & {78.3 82.0} & {48.3 49.2} & {1.7 2.9}   & {35.9 32.3} & {71.4 61.5} & {3.1 2.8}   & {42.2 45.8} & {64.1 65.3}  \\
			{EmoBERTa}                 & \textbf{78.9 82.5} & {50.2 50.2} & {1.8 1.9}   & {33.3 31.2} & {72.1} 61.7 & {9.1 2.5}   & {43.3 46.4} & {64.1 65.2}  \\
			ConGCN                       & 46.8 45.4 & 10.6 8.8 & 8.7 8.1 & 53.1 54.6 & 76.7 75.2 & 28.5 \textbf{26.3} & 50.3 48.4 & 59.4 58.7 \\
			{A-DMN }                   & {76.5 78.9} & \textbf{56.2 55.3} & {8.2 8.6}   & {22.1 24.9} & {59.8 57.4} & {1.2 3.4}   & {41.3 40.9} & {61.5 60.4}  \\
			{LR-GCN}                   & {76.7 80.0} & {53.3 55.2} & {0.0 0.0}   & {49.6 35.1} & 68.0 {64.4} & {10.7 2.7}  & {48.0 51.0} & {65.7 65.6}  \\
			DER-GCN                  & 76.8 80.6 & 50.5 51.0 & {14.8 10.4} & {56.7 41.5} & 69.3 64.3 & {17.2 10.3} & \textbf{52.5} 57.4 & {66.8 66.1}  \\
			ELR-GCN  &  \textbf{80.2  83.6} &  36.8 35.4     &  \textbf{19.2 13.1}       &  \textbf{80.2  83.6}   &   \textbf{76.5 69.7}  &  \textbf{55.6}  13.0   & 52.1 \textbf{57.7}  & \textbf{68.7 69.9} \\ \hline
	\end{tabular}}
\end{table*}

\textbf{ConGCN} \cite{zhang2019modeling} treats multimodal features as node features in the graph and utilizes heterogeneous graphs to model conversational relationship information between spe-akers.

\textbf{LR-GCN} \cite{ren2021lr} captures the latent dependencies between contexts by constructing multiple graphs and constructs densely connected layers to extract speaker relationship information and structural information of the graph.

\textbf{AGHMN} \cite{jiao2020real} uses BiGRU to fuse the correlation information between historical contexts and uses the attention mechanism to give higher weight to important context information.

\textbf{BiERU} \cite{li2022bieru} uses emotion recurrent units and emotion feature extractors to extract contextual semantic information respectively.
and refine contextual emotion feature vectors.

\textbf{EmoBERTa} \cite{kim2021emoberta} uses RoBERTa to extract sequential contextual semantic information from text. This method does not use multi-modal data.

\textbf{LFM} \cite{liu-etal-2018-efficient-low} uses low-rank decomposition to effectively reduce the dimensionality disaster problem that occurs during the fusion process of multi-modal features.

\textbf{RGAT} \cite{ishiwatari2020relation} integrates position encoding information into the graph attention network to improve the model's context understanding ability.

\textbf{CoMPM} \cite{lee2022compm} uses a pre-trained model to extract pre-trained context memory information and combines it with the context model to understand the global contextual emotional features in a fine-grained manner.

\textbf{COGMEN} \cite{joshi2022cogmen} improves the representation ability of emotional feature vectors by building context GCN to extract global and local context information and fuse them.

\textbf{DER-GCN} \cite{ai2023gcn} improves the model's emotional representation capabilities by constructing speaker relationship graphs and event graphs.

\textbf{A-DMN} \cite{xing2020adapted} A-DMN comprehensively considers the intra-speaker and inter-speaker contextual information, and uses GRU to achieve cross-modal feature fusion.

\textbf{CTNet} \cite{lian2021ctnet} realizes semantic information interaction within and between modalities by building Single Transformer and Cross Transformer.

\begin{figure*}
	\centering
	\includegraphics[width=1\linewidth]{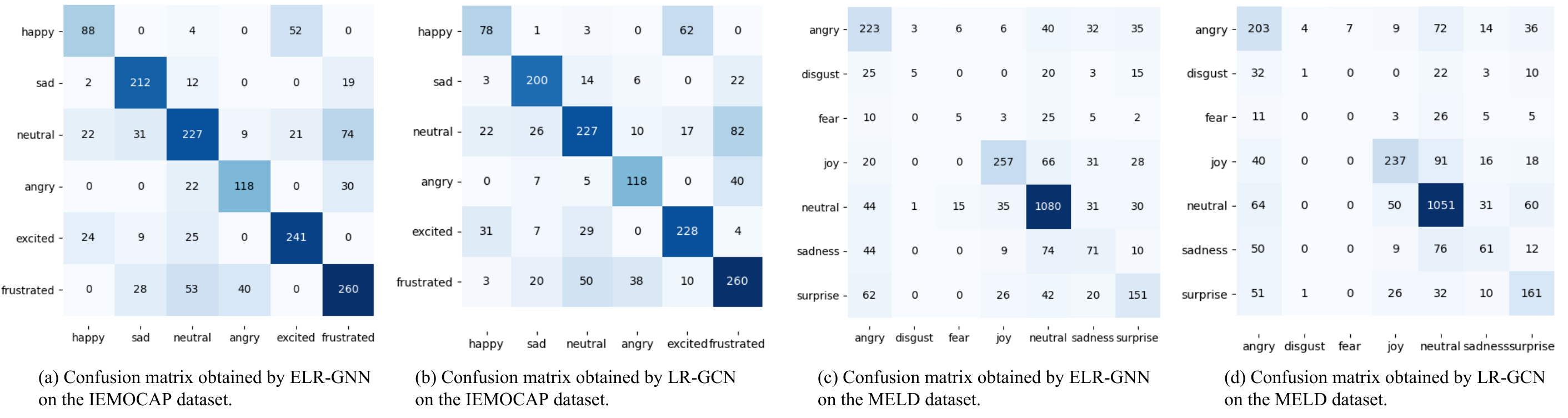}
	\caption{Confusion matrix of ELR-GNN and LR-GNN classification on IEMOCAP and MELD datasets.}
	\label{fig:confusing-matrix}
\end{figure*}

\subsection{Comparison with the State-of-the-Art Methods}
To verify the superiority of the ELR-GNN method proposed in this paper, we report the experimental results of ELR-GNN and other comparative methods on the IEMOCAP and MELD data sets. Experimental results are presented in Tables \ref{tab:iemocap} and \ref{tab:meld}.

\textbf{IEMOCAP:} As shown in Table \ref{tab:iemocap}, the multi-modal emotion recognition method proposed in this paper achie-ved the best emotion recognition effect on the IEMOCAP data set, with an average accuracy of 70.6\% and an average F1 value of 70.9\%. ELR-GCN proposes an effective modeling method of long-distance context latent dependencies for multi-modal emotion recognition. In addition, ELR-GCN also combines early and adaptive late fusion methods to achieve the capture of fine-grained emotional features. Among other comparison methods, the emotion recognition effect of DER-GCN is slightly lower than that of ELR-GNN, with an average accuracy of 69.7\% and an average F1 value of 69.4\%. Although DER-GCN comprehensively considers event relationships and dialogue relationships between speakers to enhance the model's emotional understanding, it ignores latent context dependencies. The emotion recognition effect of LR-GCN is lower than ELR-GNN and DER-GCN, with an average accuracy of 68.5\% and an average F1 value of 68.3\%. Although LR-GCN considers latent dependencies between contexts, due to the high computational complexity of GCN, LR-GCN can only capture local latent dependencies. The emotion recognition effects of other comparison methods are lower than ELR-GNN. Likewise, none of them take into account potential dependencies on context. Overall, the accuracy of ELR-GNN on the happy emotion analogy is much higher than that of other comparison algorithms, while the accuracy of other emotion categories is also relatively close to that of other comparison algorithms.  In addition, the F1 value of ELR-GNN on the happy and excited emotional analogies is much higher than that of other comparison algorithms. At the same time, the F1 value of ELR-GNN on other emotional categories is also relatively close to other comparison algorithms. The experimental results prove the superiority of the ELR-GNN method proposed in this paper.

\textbf{MELD:} As shown in Table \ref{tab:meld}, The ELR-GNN method proposed in this article has the best emotion recognition effect on the MELD data set, with an average accuracy of 68.7\% and an average F1 value of 69.9\%. The emotion recognition effect of DER-GCN is second, with an average accuracy of 69.7\% and an average F1 value of 69.4\%. The emotion recognition effect of LR-GCN is lower than that of ELR-GNN and DER-GCN, with an average accuracy of 68.5\% and an average F1 value of 68.3\%. The emotion recognition effects of other comparison methods are relatively poor, and the average accuracy and F1 value are lower than ELR-GNN. The performance improvement may be attributed to ELR-GNN's ability to capture long-distance contextual latent dependencies and fine-grained fusion of dialogue relationships between speakers, contextual latent dependencies and contextual semantic information. Overall, the accuracy of ELR-GNN on the neutral, fear, sadness, joy, and disgust emotion analogy is much higher than that of other comparison algorithms, while the accuracy of other emotion categories is also relatively close to that of other comparison algorithms.  In addition, the F1 value of ELR-GNN on the neutral, fear, sadness, joy, and anger emotional analogies is much higher than that of other comparison algorithms. At the same time, the F1 value of ELR-GNN on other emotional categories is also relatively close to other comparison algorithms. In addition, we find that ELR-GNN has better emotion recognition effects on the minority emotions fear and disgust, with relatively high accuracy and F1 value. The experimental results prove the superiority of the ELR-GNN method proposed in this paper.

\begin{figure*}
	\centering
	\includegraphics[width=1\linewidth]{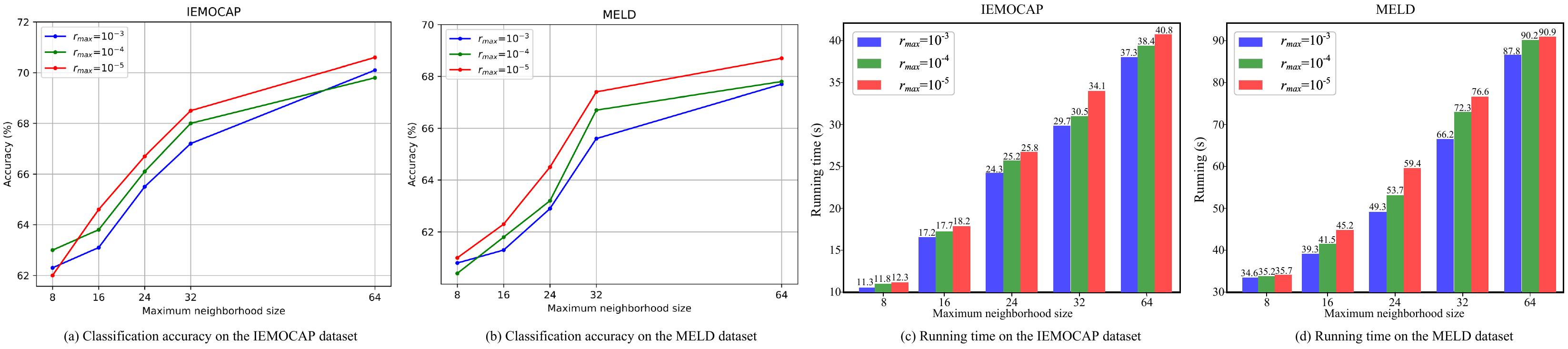}
	\caption{We tested the impact of the maximum neighborhood size and parameter $r_{max}$ in ELR-GNN on the accuracy and running time of emotion recognition.}
	\label{fig:running-time}
\end{figure*}

In addition, to intuitively illustrate that the running time of the ELR-GNN method proposed in this paper is better than other comparative methods, we statistics in Table \ref{fig:running-time} the running time of other comparative methods of the ELR-GNN method on the IEMOCAP and MELD data sets. As shown in Table \ref{fig:running-time}, the running time of the ELR-GNN method proposed in this paper on the IEMOCAP and MELD data sets is 41s and 91s respectively, which is significantly better than other comparison methods. The running times of DialogueGCN are 58s and 127s respectively, which are lower than LR-GCN and DER-GCN, but the emotion recognition effect is relatively poor. The running times of LR-GCN are 87s and 142s respectively. The running times of DER-GCN are 125s and 189s respectively. The experimental results prove the efficiency and effectiveness of the ELR-GNN method proposed in this paper.

\begin{table}[htbp]
	\caption{We tested the running time of the ELR-GNN method proposed in this paper and other comparative methods on the IEMOCAP and MELD data sets. In particular, ELR-GNN sets $r_{max}$ to $10^{-5}$ and neighbor size to 64.}
	\label{tab:running}
	\renewcommand\arraystretch{1.5}
	\setlength{\tabcolsep}{6.5mm}{
		\begin{tabular}{c|cc}
			\hline
			\multirow{2}{*}{Methods} & \multicolumn{2}{c}{Running time (s)} \\ \cline{2-3}
			& IEMOCAP            & MELD            \\ \hline
			DialogueGCN              & 58                 & 127             \\
			LR-GCN                   & 87                 & 142             \\
			DER-GCN                  & 125                & 189             \\
			ELR-GNN                  & 41               & 91            \\ \hline
	\end{tabular}}
\end{table}

\subsection{Analysis of the Experimental Results}
To intuitively understand the ability of the feeling model for each emotion category, we analyzed the emotion classification of ELR-GNN and LR-GCN on the test set. Fig. \ref{fig:confusing-matrix} shows the confusion matrix of ELR-GNN and LR-GCN for emotion classification on IEMOCAP and MELD data sets.

Overall, on the IEMOCAP data set, ELR-GCN has a higher number of correct classifications for each emotion category than LR-GCN. On the MELD dataset, ELR-GCN has a higher number of correct classifications than LR-GCN in most emotional categories. The performance improvement may be attributed to ELR-GNN's ability to understand the semantic representation of each emotion category in a fine-grained manner.

On the IEMOCAP dataset, the confusion matrix shows that ELR-GNN easily misclassifies happy emotions into excited emotions. Similarly, LR-GCN also easily misclassifies happy emotions into excited emotions, and even the number of misclassifications is greater than that of ELR-GNN. We speculate that this is because the semantics of happy emotions and excited emotions are relatively similar, and the model cannot differentiate between these two types of emotions in a fine-grained manner. In addition, we also find that ELR-GNN easily misclassifies neutral emotions into frustated emotions.

On the MELD data set, the confusion matrix shows that ELR-GNN has a very poor classification effect on disgust and fear emotions, and can only correctly classify a few samples. This is because the number of disgust and fear emotion categories is relatively small, and the data set has a serious imbalance problem, which leads to deviations in the model's emotional understanding ability. The number of correct classifications of ELR-GNN on neutral emotions is very large, and there are very few misclassified samples. Experimental results prove that ELR-GNN has a relatively strong ability to understand neutral emotional categories.

\subsection{Ablation Study}

\subsubsection{Importance of the Modalities}
To verify the importance of the three modal features of text, video and audio for ELR-GNN, we conducted ablation experiments on the IEMOCAP and MELD data sets to compare the performance of the combination of different modal features. The experimental results are shown in Table \ref{tab:multi-modal}. In single-modal experiments, ELR-GNN with text modality features has the best emotion recognition effect. The average accuracy on the IEMOCAP and MELD data sets are 64.1\% and 63.5\%, respectively, and the average F1 value is 63.9\% and 62.4\%, respectively. The emotion recognition effect of ELR-GNN with audio modal features is second, with average accuracy rates of 61.1\% and 62.7\% on the IEMOCAP and MELD data sets, and average F1 values of 60.8\% and 62.0\% respectively. ELR-GNN with video modality features has the worst emotion recognition effect, with average accuracy rates of 59.4\% and 60.1\% on the IEMOCAP and MELD data sets, and average F1 values of 59.7\% and 61.4\% respectively. Experimental results show that text features contain the most emotional semantic information. In the dual-modal experiment, ELR-GNN with text and audio modal features has the best emotion recognition effect. The average accuracy on the IEMOCAP and MELD data sets are 65.0\% and 64.1\%, respectively, and the average F1 values are are 64.4\% and 63.2\%, respectively. Experimental results demonstrate the effectiveness of multimodal features.

\begin{table}[htbp]
	\renewcommand\arraystretch{1.5}
	\setlength{\tabcolsep}{4.5mm}{
		\caption{The effect of ELR-GNN on IEMOCAP and MELD datasets using unimodal features and multimodal features, respectively. We report average accuracy and F1-score.}
		\label{tab:multi-modal}
		\begin{tabular}{c|cccc}
			\hline
			\multirow{2}{*}{Modality} & \multicolumn{2}{c}{IEMOCAP} & \multicolumn{2}{c}{MELD} \\ \cline{2-5}
			& Acc.       & F1              & Acc         & F1         \\ \hline
			T                         & 64.1        & 63.9       & 63.5          & 62.4           \\
			A                         & 61.1        & 60.8                & 62.7          & 62.0           \\
			V                       & 59.4        & 59.7                & 60.1          & 61.4
			\\
			T+A                       & 65.0        & 64.4                & 64.1          & 63.2           \\
			T+V                       & 64.3        & 64.6                & 64.0          & 62.9           \\
			V+A                       & 63.0        & 62.7                & 61.5          & 61.9           \\
			T+A+V                     &\textbf{70.6}      & \textbf{70.9}               &    \textbf{68.7}        &  \textbf{69.9}        \\ \hline
	\end{tabular}}
\end{table}

\subsubsection{Parameter Analysis}
We tested the impact of the maximum neighborhood size and parameter $r_{max}$ in ELR-GNN on the accuracy and running time of emotion recognition. As shown in Figs. \ref{fig:running-time}(a), and \ref{fig:running-time}(b), we tested the impact of different neighborhood sizes and $r_{max}$ on emotion recognition accuracy on the IEMOCAP and MELD datasets. Experimental results show that when $r=10^{-5}$, ELR-GNN has the best emotion recognition effect. When $r=10^{-4}$, the emotion recognition effect of ELR-GNN is second. When $r=10^{-3}$, ELR-GNN has the worst emotion recognition effect. Furthermore, as the size of the neighborhood continues to increase, the model's emotion recognition performance also improves. Experimental results demonstrate the necessity of capturing long-range latent context dependencies.

 As shown in Figs. \ref{fig:running-time}(c), and \ref{fig:running-time}(d), We also calculated the impact of different neighborhood sizes on running time and emotion recognition accuracy. Experimental results show that as the neighborhood size increases, the running time of the model also increases, but it is lower than the running time of LR-GCN and DER-GCN. In addition, as the neighborhood size increases, the emotion recognition effect of the model also improves.

\section{Conclusions}
In this paper, we propose a novel Efficient Long-distance Latent Relation-aware Graph Neural Network (ELR-GNN) for multi-modal emotion recognition. Specifically, we first use RoBERTa, 3D-CNN and openSMILE to perform pre-feature extraction of text, video and audio features respectively. Next, we use Bi-LSTM to capture contextual semantic information and obtain low-level utterence features. We then use low-level utterence features to construct a speaker graph. In the constructed speaker relationship graph, low-level utterence features are used as node features, while dialogue relationship information between speakers is used for edge construction. To capture the latent dependency information between long-distance contexts, we use the graph random neural network algorithm to randomly sample top-$k$ nodes for information extraction. In addition, we combine early fusion and adaptive late fusion mechanisms to simultaneously fuse speaker relationship information and latent dependency information between contexts. On the IEMOCAP and MELD data sets, the ELR-GNN method proposed in this paper is better than other comparative methods, and the experimental results prove the superiority of the ELR-GNN method.

\section*{CRediT authorship contribution statement} \label{contribution}
\textbf{Yuntao Shou}: Conceptualization, Methodology, Software, Data curation,Visualization, Validation, Writing - original draft, Writing - review \& editing.
\textbf{Wei Ai}: Conceptualization, Methodology, Writing - review \& editing.
\textbf{Jiayi Du}: Conceptualization, Methodology, Writing - review \& editing.
\textbf{Haiyan Liu}: Conceptualization, Methodology, Writing - review \& editing.
\textbf{Tao Meng}: Conceptualization, Methodology, Software, Visualization, Validation, Writing - original draft, Writing - review \& editing, Funding acquisition.

\section*{Data availability} \label{data availability}
Data will be made available on request.

\section*{Declaration of competing interest} \label{Declaration}
The authors declare that they have no known competing financial interests or personal relationships that could have appeared to influence the work reported in this paper.

\section*{Acknowledgments} \label{Acknowledgments}
This work is supported by National Natural Science Foundation of China (Grant No. 61802444), the Research Foundation of Education Bureau of Hunan Province of China (GrantNo. 22B0275).


\bibliographystyle{elsarticle-num}
\bibliography{refs}


%
%
%

\end{document}